\documentclass[runningheads]{llncs}
\usepackage[T1]{fontenc}
\usepackage{graphicx}
\usepackage{utils}

\begin{document}
\title{Natural Gradient Interpretation of\\Rank-One Update in CMA-ES}
\author{
Ryoki Hamano\inst{1}%\orcidID{0000-0002-4425-1683}
\and
Shinichi Shirakawa\inst{2}%\orcidID{0000-0002-4659-6108}
\and
Masahiro Nomura\inst{1}%\orcidID{0000-0002-4945-5984}
}
\authorrunning{Hamano et al.}
\institute{CyberAgent, Inc., Shibuya, Japan\\
\email{
hamano\_ryoki\_xa@cyberagent.co.jp,
nomura\_masahiro@cyberagent.co.jp,
}
\and
Yokohama National University, Yokohama, Japan\\
\email{
shirakawa-shinichi-bg@ynu.ac.jp
}
}
\maketitle              % typeset the header of the contribution
\begin{abstract}
% 220 words
The covariance matrix adaptation evolution strategy (CMA-ES) is a stochastic search algorithm using a multivariate normal distribution for continuous black-box optimization.
In addition to strong empirical results, part of the CMA-ES can be described by a stochastic natural gradient method and can be derived from information geometric optimization (IGO) framework.
However, there are some components of the CMA-ES, such as the rank-one update, for which the theoretical understanding is limited.
While the rank-one update makes the covariance matrix to increase the likelihood of generating a solution in the direction of the evolution path, this idea has been difficult to formulate and interpret as a natural gradient method unlike the rank-$\mu$ update.
In this work, we provide a new interpretation of the rank-one update in the CMA-ES from the perspective of the natural gradient with prior distribution.
First, we propose maximum a posteriori IGO (MAP-IGO), which is the IGO framework extended to incorporate a prior distribution.
Then, we derive the rank-one update from the MAP-IGO by setting the prior distribution based on the idea that the promising mean vector should exist in the direction of the evolution path.
Moreover, the newly derived rank-one update is extensible, where an additional term appears in the update for the mean vector.
We empirically investigate the properties of the additional term using various benchmark functions.
% Moreover, the newly derived rank-one update is extensible and its properties are empirically investigated using various benchmark functions.
% Moreover, the rank-one update newly derived is extensible and experimental results suggest that the extended rank-one update improves the performance of the CMA-ES on various benchmark functions.

\end{abstract}
%
%
%
% In addition to strong empirical results, the update of the distribution parameters in the CMA-ES can be partially described by the stochastic natural gradient ascent.
% In this work, we propose a new principled framework, maximum a posteriori CMA-ES (MAP-CMA), in which knowledge such as rank-one update can be integrated into the CMA-ES as a prior information.
% We derive the rank-one update from the MAP-CMA with the prior information that the mean vector is more likely to exist in the direction of the evolution path.

% ========================================
\section{Introduction} \label{sec:intro}
% ========================================
The covariance matrix adaptation evolution strategy (CMA-ES)~\cite{hansen_reducing_2003,hansen_adapting_1996,hansen2016cma} is recognized as a state-of-the-art derivative-free stochastic algorithm for black-box continuous optimization problems.
The CMA-ES proceeds the optimization by repeatedly sampling from the multivariate normal distribution and updating the distribution parameters such as the mean vector, the covariance matrix, and the step-size.
Despite the small number of hyperparameters to be tuned, the CMA-ES shows high performance for non-linear, ill-conditioned, or multimodal problems~\cite{rios2013derivative,hansen2004evaluating}.

In addition to its empirical success, the CMA-ES has attractive theoretical properties, such as its relationship to the natural gradient method.
The update of the distribution parameters in the CMA-ES can be partially described by the stochastic natural gradient ascent~\cite{Akimoto:2010}.
In fact, the pure rank-$\mu$ update CMA-ES is an instance of information geometric optimization (IGO) \cite{IGO:2017}, a unified framework for probabilistic model-based black-box optimization algorithms.

However, the theoretical understanding of the other components of the CMA-ES, such as the rank-one update, is limited.
% In the rank-one update, the covariance matrix is updated to increase the likelihood of generating a solution in the direction of the evolution path.
The rank-one update makes the covariance matrix to increase the likelihood of generating a solution in the direction of the evolution path, which accumulates the updating direction of the mean vector.
% Li and Zhang~\cite{Evopath:2016} showed that the evolution path acts as a \emph{momentum} in the rank-one update.
Li and Zhang~\cite{Evopath:2016} discuss that the evolution path cancels opposite update directions of the mean vector and the rank-one update term with the evolution path serves as momentum term for the covariance matrix.
% Then, the rank-one update realizes that the covariance matrix is updated so that candidate solutions are more likely to be generated in the direction the mean vector moves.
However, unlike the rank-$\mu$ update, these ideas of the rank-one update have been difficult to formulate and interpret based on natural gradients.

In this work, we provide a new interpretation of the rank-one update in the CMA-ES from the perspective of the natural gradient with prior distribution. 
To this end, we first propose maximum a posterioiri IGO (MAP-IGO), which is the IGO framework extended to incorporate a prior distribution.
The MAP estimation approach has provided new interpretations or extensions to existing methods, such as regularization in linear regression~\cite{bishop2006pattern} and reinforcement learning~\cite{MPO:2018,VMPO:2020}.
Then, we derive the rank-one update from the MAP-IGO with the prior distribution set based on the idea that the promising mean vector should exist in the direction of the evolution path.
Moreover, the newly derived rank-one update is extensible, where an additional term, we call \emph{momentum update}, appears in the update for the mean vector depending on the setting of the prior distribution.
We propose the CMA-ES with the momentum update as maximum a posteriori CMA-ES (MAP-CMA) and empirically investigate its properties using various benchmark functions.
% Numerical experiments show that the MAP-CMA outperforms the CMA-ES on various benchmark functions.

% In this work, we propose a new principled framework, maximum a posteriori CMA-ES (MAP-CMA), in which knowledge such as rank-one update can be integrated into the CMA-ES as a prior information.
% The MAP estimation approach has provided new interpretations or extensions to existing methods, such as regularization in linear regression and reinforcement learning~\cite{VMPO:2020}.
% We derive the rank-one update from the MAP-CMA with the prior information that the mean vector is more likely to exist in the direction of the evolution path.
% Moreover, the rank-one update newly derived is extensible, in which an additional term appears in the update for the mean vector depending on the setting of the prior distribution.
% Numerical experiments show that the extended rank-one update improves performance of the CMA-ES on various benchmark functions.

The remainder of this paper is organized as follows.
Section~\ref{sec:background} shows background of this study, which is needed in the rest of the paper.
In Section~\ref{sec:mapigo}, we propose the MAP-IGO and derive the update rule when the multivariate normal distribution and the normal-inverse-Wishart distribution are applied to the MAP-IGO.
In Section~\ref{sec:rankone}, we provide a new interpretation of the rank-one update in the CMA-ES and propose the MAP-CMA.
Section~\ref{sec:experiments} shows experimental results for the MAP-CMA on benchmark functions.
Section~\ref{sec:conclusion} concludes the paper with future work.

% Our contributions are summarized as follows:
% \begin{itemize}
% \item We propose a new principled framework, \emph{MAP-CMA}, in which prior information can be integrated into the CMA-ES.
% \item By setting the prior information of the mean vector based on the evolution path, the rank-one update in the CMA-ES can be derived.
% \item The rank-one update derived from setting the prior information can be extended and contributes to better performance than the original rank-one update.
% \end{itemize}

% The rest of this paper is organized as follows.
% Section~\ref{sec:background} shows background of this study, which is needed in the remaining paper.
% In Section~\ref{sec:mapigo}, we extend the existing IGO framework to incorporate prior distribution in a principled way.
% In Section~\ref{sec:mapcma}, we propose a novel method, MAP-CMA, based on the extended IGO framework.
% In Section~\ref{sec:rankone}, we provide an interpretation of the rank-one update with the prior information.
% Section~\ref{sec:experiments} shows experimental results for the MAP-CMA with newly derived rank-one update.
% Section~\ref{sec:conclusion} concludes this study and discuss the future work. % \label{sec:intro}
% ========================================
\section{Preliminaries} \label{sec:background}
% ========================================

% ----------------------------------------
\subsection{CMA-ES}
% ----------------------------------------
The covariance matrix adaptation evolution strategy (CMA-ES) generates multiple solutions in each iteration from the multivariate normal distribution $\mathcal{N}(\m, \linebreak (\sig[t])^2 \C)$.
To generate candidate solutions from the distribution, the CMA-ES updates the distribution parameter, the mean vector $\m \in \mathbb{R}^N$, and the covariance matrix $(\sig[t])^2 \C \in \mathbb{R}^{N \times N}$.
This update partially corresponds to the natural gradient ascent in the parameter space, which is the steepest ascent with respect to the Fisher metric~\cite{amari2000methods}.
We will describe \emph{information geometric optimization}~\cite{IGO:2017}, a framework that generalizes this natural gradient ascent step, in Section~\ref{ssec:igo}.
This section introduces the well-known CMA-ES variant, $(\mu/\muw, \lambda)$-CMA-ES, minimizing the objective function $f(\x)$.

In the $t$-th iteration, the $\lambda$ candidate solutions $\x_i$ ($i=1, 
\ldots,\lambda$) are generated as
\begin{align}
    \y_i &\sim \mathcal{N}(\boldsymbol{0}, \C[t]) \enspace, \label{eq:sample_x} \\
    \x_i &= \m[t] + \sig[t] \y_i \enspace. \label{eq:sample_y}
\end{align}
%After evaluating the candidate solutions with the objective function, we sort $\{\x_{1:\lambda}, \x_{2:\lambda}, \ldots, \x_{\lambda:\lambda}\}$ and $\{\y_{1:\lambda}, \y_{2:\lambda}, \ldots, \y_{\lambda:\lambda}\}$ so that the indices correspond to $f(\x_{1:\lambda}) \leq f(\x_{2:\lambda}) \leq \ldots \leq f(\x_{\lambda:\lambda})$.

The mean vector $\m[t]$ is updated as
\begin{align}
    \m[t+1] = \m[t] + c_m \sum_{i=1}^{\lambda} w_{i} ( \x_{i:\lambda} - \m[t] ) \enspace, \label{eq:cmaes-m}
\end{align}
where $c_m$ is the learning rate for the mean vector, and $w_i$ is the weight that holds $\sum_{i=1}^{\mu} \w_i = 1, \w_1 \geq \w_2 \geq \cdots \geq \w_{\mu} > 0$, $w_{j} = 0\ (j = \mu+1, \ldots,\lambda)$.
The index of the $i$-th best sample is denoted as $i\!:\!\lambda$.

The CMA-ES employs the two evolution paths\footnote{The CMA-ES sometimes employs the indicator function $h_\sigma$ to prevent evolution path $\pc[t]$ from rapidly lengthening.}.
\begin{align}
    \ps[t+1] &= (1-c_\sigma)\ps[t] + \sqrt{c_\sigma(2-c_\sigma)\muw} {\C[t]}^{-\frac12} \sum_{i=1}^\lambda \w_i \y_{i:\lambda} \enspace, \\
    \pc[t+1] &= (1-c_c) \pc[t] + \sqrt{c_c(2-c_c)\muw} \sum_{i=1}^\lambda \w_i \y_{i:\lambda} \enspace,
\end{align}
where $\muw = 1 / \sum_{i=1}^{\mu} \w_i^2$, $c_\sigma$ and $c_c$ are cumulative rates. %, and
%\begin{align*}
%    h_\sigma \leftarrow \mathds{1} {\left\{\|\ps[t+1]\| < \sqrt{1-(1-c_\sigma)^{2(t+1)}}\left(1.4+\frac{2}{N+1}\right)\E [\|\mathcal{N}(\boldsymbol{0}, \mathbf{I}) \|]\right\}}
%\end{align*}
%is the indicator function to prevent evolutionary path $\pc[t]$ from rapidly lengthening.
% The expected norm $\E [\|\mathcal{N}(\boldsymbol{0}, \mathbf{I}) \|]$ is practically approximated by $\sqrt{N} \left( 1 - \frac{1}{4 N} + \frac{1}{21 N^2} \right)$.

The covariance matrix $\C[t]$ is updated as
\begin{align}
    % \C[t+1] = \left( 1-c_1-c_{\mu} \sum_{i=1}^{\lambda}  \right)
    &\C[t+1] \leftarrow \C[t] + \underbrace{c_1 \left(\pc[t+1]{\pc[t+1]}^\top - \C[t] \right) }_{\text{rank-one update}} + \underbrace{c_\mu \sum_{i=1}^{\lambda} w_i \left( \y_{i:\lambda}\y_{i:\lambda}^\top - \C[t] \right)}_{\text{rank-}\mu\text{ update}} \enspace, \label{eq:cmaes-C}
\end{align}
where $c_1$ and $c_{\mu}$ are the learning rates for the rank-one update and the rank-$\mu$ update, respectively.

The step-size $\sig[t]$ is updated as
\begin{align}
    \sig[t+1] \leftarrow \sig[t] \exp \left( \frac{c_\sigma}{d_\sigma} \left( \frac{\|\ps[t+1] \|}{\E [\|\mathcal{N}(\boldsymbol{0}, \mathbf{I}) \|]} - 1 \right) \right) \enspace,
\end{align}
where $d_\sigma$ is a damping factor.
The expected norm $\E [\|\mathcal{N}(\boldsymbol{0}, \mathbf{I}) \|]$ is practically approximated by $\sqrt{N} \left( 1 - \frac{1}{4 N} + \frac{1}{21 N^2} \right)$.
The CMA-ES that only employs the mean vector update and the rank-$\mu$ update is called \emph{pure rank-$\mu$ update CMA-ES}~\cite{hansen_reducing_2003}.

% ----------------------------------------
\subsection{Information Geometric Optimization} \label{ssec:igo}
% ----------------------------------------
Information geometric optimization~(IGO)~\cite{IGO:2017} is a unified framework of probabilistic model-based black-box optimization algorithms. Given a family of probability distributions $\{P_{\btheta} \}$ on $X$ parameterized by $\btheta \in \Theta$, the IGO transforms the original problem into the maximization of the expected value of $J_{\btheta^{(t)}}: \Theta \to \R$. The function $J_{\btheta^{(t)}}$ depends on the current distribution parameter $\btheta^{(t)}$ and is defined as the expectation of the utility function $W_{\btheta^{(t)}}^f (\x)$ over $p_{\btheta} (\x)$, i.e.,
\begin{align}
    J_{\btheta^{(t)}} (\btheta) = \int W_{\btheta^{(t)}}^f (\x) p_{\btheta} (\x) \diff \x \enspace. \label{eq:igo_argmax}
\end{align}
The utility function $W_{\btheta^{(t)}}^f (\x)$ is defined based on the quantiles of $f$ under the current distribution. This approach provides invariance under increasing transformations of the objective function. Let $q_{\btheta}^\leqslant (\x) = \Pr_{\x' \sim p_{\btheta}} (f(\x') \leq f(\x))$ and $w : [0, 1] \to \R$ be a non-increasing function. Assuming that the probability of having the same evaluation value for different samples is $0$, the utility function $W_{\btheta^{(t)}}^f (\x)$ is defined as $w(q_{\btheta^{(t)}}^\leqslant (\x))$. See \cite{IGO:2017} for the definition of $W_{\btheta^{(t)}}^f (\x)$ when this assumption is not satisfied.

The IGO maximizes Eq.~\eqref{eq:igo_argmax} by natural gradient ascent. The natural gradient $\ngr_\btheta J_{\btheta^{(t)}} (\btheta)$ is given by the product of the inverse of Fisher information matrix and the vanilla gradient, namely, $F^{-1}(\btheta) \nabla_{\btheta} J_{\btheta^{(t)}} (\btheta)$. Then, the natural gradient $\ngr_{\btheta} J_{\btheta^{(t)}} (\btheta)$ is calculated as follows:
\begin{align}
    \ngr_{\btheta} J_{\btheta^{(t)}} (\btheta) = \int W_{\btheta^{(t)}}^f (\x) \ngr_{\btheta} (\ln p_{\btheta} (\x)) p_{\btheta} (\x)  \diff \x \enspace. \label{eq:IGO-ngr}
\end{align}
In the black-box setting, Eq.~\eqref{eq:IGO-ngr} cannot be computed analytically, hence the natural gradient of $J_{\btheta^{(t)}} (\btheta)$ at $\btheta = \btheta^{(t)}$ is approximated by the Monte Carlo estimation using $\lambda$ samples $\x_{1}, \ldots, \x_{\lambda}$ generated from $P_{\btheta^{(t)}}$ as
\begin{align}
    \left. \ngr_{\btheta} J_{\btheta^{(t)}} (\btheta) \right|_{\btheta = \btheta^{(t)}} \approx \frac{1}{\lambda} \sum_{i=1}^\lambda \hat{w}_i \left. \ngr_{\btheta} \ln p_{\btheta} (\x_{i:\lambda}) \right|_{\btheta = \btheta^{(t)}} \enspace,
\end{align}
where $\hat{w}_i$ is the ranking-based utility value that can be regarded as the estimation value of $W_{\btheta^{(t)}}^f (\x_i)$. %and $i\!:\!\lambda$ denotes the index of the $i$-th best sample.
Introducing the learning rate $\eta$, we obtain the update rule.
\begin{align}
    \btheta^{(t+1)} = \btheta^{(t)} + \eta  \sum_{i=1}^\lambda \frac{\hat{w}_i}{\lambda} \left. \ngr_{\btheta} \ln p_{\btheta} (\x_{i:\lambda}) \right|_{\btheta = \btheta^{(t)}} \enspace.
\end{align}
The IGO framework recovers the pure rank-$\mu$ update CMA-ES when the family of normal distributions is applied, and the population-based incremental learning~(PBIL)~\cite{PBIL:1994} when the family of Bernoulli distributions is applied.

% \label{sec:background}
% ========================================
\section{Maximum a Posteriori IGO} \label{sec:mapigo}
% ========================================
In this section, we propose maximum a posteriori IGO (MAP-IGO), which is the IGO framework extended to incorporate a prior distribution in a principled way.
Furthermore, we apply the multivariate normal distribution and the normal-inverse-Wishart distribution, which is the conjugate prior distribution of the multivariate normal distribution, to the MAP-IGO.

\subsection{Introducing Prior Information to IGO}
% In this section, we extend the existing IGO framework to reflect prior information.
% We assume that we have information about promising distribution parameters.
% To make use of this information, we first rewrite the existing IGO objective in the form of maximum likelihood (ML) estimation.
First, we rewrite the existing IGO objective in the form of maximum likelihood (ML) estimation.
Then, we further rewrite the ML estimation in the form of the maximum a posteriori (MAP) estimation.
% By setting the prior distribution in the MAP estimation using the promising distribution parameters, we can reflect the prior information in a principled way.
% Specifically, this extension consists of the following two phases:
% (i) Given a specific definition of a binary outcome, we first show that maximizing the existing IGO objective is equivalent to maximum likelihood (ML) estimation of the binary outcome (i.e. inference problem).
% (ii) By integrating prior distribution of the distribution parameters as prior information, we extend the maximum likelihood estimation to the maximunm a posteriori (MAP) estimation.
% We will describe each of these phases in detail below.

%\noindent
%{\bf Equivalence of IGO objective to ML estimation}\\
\paragraph{Equivalence of IGO objective to ML estimation}
%We first rewrite the existing IGO objective in the form of maximum likelihood (ML) estimation.
To rewrite the existing IGO objective in the form of the ML estimation, we introduce a binary event $R_{\btheta^{(t)}} \in \{ 0, 1 \}$ where it holds $p(R_{\btheta^{(t)}}=1 | \x) \propto W_{\btheta^{(t)}}^f (\x)$, inspired by~\cite{TR-CMA-ES:2017,MPO:2018,VMPO:2020}.
It should be noted that $R_{\btheta^{(t)}}=1$ depends on the current distribution parameter $\btheta^{(t)}$.
Intuitively, the probability $p(R_{\btheta^{(t)}}=1 | \x) \propto W_{\btheta^{(t)}}^f (\x)$ is larger when $\x$ has a better objective function value than solutions sampled from $p_{\btheta^{(t)}}$.
Here, we consider the marginal distribution of the event $R_{\btheta^{(t)}}=1$ over $\btheta$.
\begin{align}
    p (R_{\btheta^{(t)}}=1 | \btheta) = \int p(R_{\btheta^{(t)}}=1 | \x) p_{\btheta}(\x) \diff \x \enspace.
\end{align}
The important thing here is that the maximization of the IGO objective can be rewritten in the form of the ML estimation as follows:
\begin{align}
    &\argmax_{\btheta \in \Theta} J_{\btheta^{(t)}} (\btheta) = \argmax_{\btheta \in \Theta} p (R_{\btheta^{(t)}} = 1 \mid \btheta) \enspace. \label{eq:equiv-IGO-MLE}
\end{align}

%\noindent
%{\bf MAP estimation instead of ML estimation}\\
\paragraph{MAP estimation instead of ML estimation}
Eq.~(\ref{eq:equiv-IGO-MLE}) allows us to consider the optimization of the IGO objective as a kind of ML estimation.
% Therefore, it is natural to consider the MAP estimation~\cite{bishop2006pattern} instead of the ML estimation, as we set the prior distribution parameters.
%, as we wish to set the prior distribution parameters for the prior distribution.
Then, we can introduce the prior distribution into the IGO by taking the approach of the MAP estimation~\cite{bishop2006pattern}.
To that end, we calculate the posterior distribution by using the Bayes' theorem as follows:
\begin{align}
    p (\btheta | R_{\btheta^{(t)}}=1) \propto p (R_{\btheta^{(t)}}=1 | \btheta) p(\btheta) \enspace,
\end{align}
where $p(\btheta)$ is the prior distribution.
We note that the prior distribution $p(\btheta)$ can be set at each time depending on the current parameter $\btheta^{(t)}$.
The resulting framework, MAP-IGO, estimates $\btheta$ as the mode of the posterior distribution:
\begin{align}
    \argmax_{\btheta \in \Theta} \ln p (\btheta \mid R_{\btheta^{(t)}} = 1) = \argmax_{\btheta \in \Theta} \ln p (R_{\btheta^{(t)}} = 1 \mid \btheta) + \ln p(\btheta) \enspace. \label{eq:MAP-IGO-argmax}
\end{align}

\subsection{Natural Gradient Update for MAP-IGO}
To optimize Eq.~\eqref{eq:MAP-IGO-argmax} using the natural gradient ascent, we calculate the natural gradient of $\ln p (R_{\btheta^{(t)}} = 1 \mid \btheta) + \ln p(\btheta)$.
We assume that $\int W_{\btheta^{(t)}}^f (\x) p_\btheta(\x) \diff \x (=\int_0^1 w(q) \diff q) \neq 0$ holds in this work\footnote{It should be noted that while the assumption $\int_0^1 w(q) \diff q \neq 0$ usually holds in the CMA-ES, some instances of IGO, such as compact genetic algorithm~\cite{cGA:1999}, do not satisfy this. We will not pursue this limitation in depth as our focus is on the CMA-ES.}.
First, the natural gradient $\ngr_{\btheta} \ln p (R_{\btheta^{(t)}} = 1 \mid \btheta)$ can be calculated as follows:
\begin{align}
    &\ngr_{\btheta} \ln p (R_{\btheta^{(t)}} = 1 \mid \btheta) \notag \\
    &\qquad = \ngr_{\btheta} \ln \int W_{\btheta^{(t)}}^f (\x) p_{\btheta}(\x) \diff \x \\
    &\qquad = \frac{1}{\int W_{\btheta^{(t)}}^f (\x) p_\btheta(\x) \diff \x} \ngr_{\btheta} \int W_{\btheta^{(t)}}^f (\x) p_\btheta(\x) \diff x \\
    &\qquad = \frac{1}{\int W_{\btheta^{(t)}}^f (x) p_\btheta(\x) \diff \x} \int W_{\btheta^{(t)}}^f (\x) \ngr_{\btheta} (\ln p_\btheta(\x)) p_\btheta(\x) \diff \x \enspace. \label{eq:MAP-IGO-ngr}
\end{align}
According to \cite{IGO:2017}, $\int W_{\btheta^{(t)}}^f (\x) p_{\btheta^{(t)}}(\x) \diff \x$ is constant and always equal to the average weight $\int_0^1 w(q) \diff q$.
%Assuming that the average weight is equal to $1$, $\int W_{\theta^{(t)}}^f (x) p_{\theta^{(t)}}(x) \diff x = \int_0^1 w(q) \diff q = 1$ and $\sum_{i=1}^\lambda \hat{w}_i = 1$.
Then, the natural gradient Eq.~\eqref{eq:MAP-IGO-ngr} at $\btheta = \btheta^{(t)}$ can be approximated by Monte Carlo estimation as follows, as with the original IGO framework:
\begin{align}
    \frac{1}{\frac{1}{\lambda} \sum_{i=1}^\lambda \hat{\w}_i} \cdot \frac{1}{\lambda} \sum_{i=1}^\lambda \hat{\w}_i \left. \ngr_\btheta \ln p_\btheta (\x_{i:\lambda}) \right|_{\btheta = \btheta^{(t)}} \enspace.
\end{align}
Note that $\int W_{\btheta^{(t)}}^f (\x) p_{\btheta^{(t)}}(\x) \diff \x = \int_0^1 w(q) \diff q \approx \frac{1}{\lambda} \sum_{i=1}^\lambda \hat{\w}_i$.
Finally, using the natural gradient of $\ln p(\btheta)$ at $\btheta = \btheta^{(t)}$, the updated parameter $\theta^{(t+1)}$ is given as follows:
\begin{align}
    \btheta^{(t+1)} = \btheta^{(t)} + \eta \left( \sum_{i=1}^\lambda \frac{\hat{\w}_i}{\sum_{j=1}^\lambda \hat{\w}_j} \left. \ngr_\btheta \ln p_\btheta (\x_{i:\lambda}) \right|_{\btheta = \btheta^{(t)}} \! + \left. \ngr_\btheta \ln p(\btheta) \right|_{\btheta = \btheta^{(t)}} \! \right) \label{eq:MAP-IGO-update}
\end{align}
When we apply the multivariate normal distribution to the probability distribution and the normal-inverse-Wishart distribution, which is the conjugate prior of the multivariate normal distribution, to the prior distribution, we can derive the pure rank-$\mu$ update CMA-ES that can incorporate prior information. In addition, when we apply the family of Bernoulli distributions to the probability distribution and the family of beta distributions, where the beta distribution is the conjugate prior of the Bernoulli distribution, to the prior distribution, we can derive the PBIL that can incorporate prior distribution.

\subsection{Natural Gradient for Normal-Inverse-Wishart Distribution}
We apply the multivariate normal distribution to $p_\btheta(\x)$ and the normal-inverse-Wishart distribution to $p(\btheta)$.
The normal-inverse-Wishart distribution is defined by
\begin{align}
    p(\btheta) = \mathcal{N} \left( \boldsymbol{m} \mid \boldsymbol{\delta}, \frac{1}{\gamma} \boldsymbol{C} \right) \mathcal{W}^{-1} ( \boldsymbol{C} \mid \PSI, \nu) \enspace.
\end{align}
In the normal-inverse-Wishart distribution, the multivariate normal distribution is given as
\begin{align*}
    \mathcal{N} \left( \boldsymbol{m} \mid \boldsymbol{\delta}, \frac{1}{\gamma} \boldsymbol{C} \right) =  \left( \frac{\gamma}{2\pi} \right)^{\frac{N}{2}} |\boldsymbol{C}|^{-\frac12} \exp \left( -\frac{\gamma}{2} (\boldsymbol{m} - \boldsymbol{\delta})^\top \boldsymbol{C}^{-1} (\boldsymbol{m} - \boldsymbol{\delta}) \right) \:,
\end{align*}
where $\boldsymbol{\delta} \in \R^N$ and $\gamma > 0$. The inverse-Wishart distribution is given as
\begin{align*}
    \mathcal{W}^{-1} ( \boldsymbol{C} \mid \PSI, \nu) = \frac{|\PSI|^{\frac{\nu}{2}}}{2^{\frac{\nu N}{2}}\Gamma_N \left( \frac{\nu}{2} \right)} |\boldsymbol{C}|^{-\frac{\nu + N + 1}{2}}  \exp\left( - \frac12 \Tr \left( \PSI \boldsymbol{C}^{-1} \right) \right) \enspace,
\end{align*}
where $\Gamma_N (\cdot)$ is the multivariate gamma function, $\nu > N-1$, and $\PSI \in \R^{N \times N}$ is a positive define matrix.

We derive the natural gradient of the log-likelihood of the normal-inverse-Wishart distribution over the parameter space of $p_\btheta (\x)$.
Let $\btheta = [\boldsymbol{m}^\top \!, \VEC (\boldsymbol{C})^\top ]^\top$\!, where $\VEC (\boldsymbol{C})$ represents the matrix $\boldsymbol{C}$ rearranged into a column vector.
% This parameterization can be found in \cite{Akimoto:2010}.
%Let $\btheta = [\boldsymbol{m}^\top, \VEC (\boldsymbol{C})^\top ]^\top$.
The vanilla gradient of the log-likelihood of $p(\btheta)$ is given as follows:
\begin{align}
    \nabla_\btheta \ln p(\btheta) = \left[
    \begin{array}{c}
        -\gamma \boldsymbol{C}^{-1} (\boldsymbol{m} - \boldsymbol{\delta}) \\[5pt]
        \begin{aligned}
            &\frac12 \VEC \left( \gamma \boldsymbol{C}^{-1} (\boldsymbol{m} - \boldsymbol{\delta}) (\boldsymbol{m} - \boldsymbol{\delta})^\top  \boldsymbol{C}^{-1} \right. \\
            &\qquad\quad \left. - (\nu + N + 2) \boldsymbol{C}^{-1} + \boldsymbol{C}^{-1} \PSI \boldsymbol{C}^{-1} \right)
        \end{aligned}
    \end{array}
    \right]
\end{align}
According to \cite{Akimoto:2010}, the Fisher information matrix $F(\btheta)$ with respect to the multivariate normal distribution parameter and its inverse matrix are, respectively,
\begin{align*}
    F(\btheta) = \left[
    \begin{array}{cc}
        \boldsymbol{C}^{-1} & \boldsymbol{0} \\
        \boldsymbol{0} & \frac12 \boldsymbol{C}^{-1} \otimes \boldsymbol{C}^{-1}
    \end{array}
    \right] \enspace \text{and} \enspace F^{-1}(\btheta) = \left[
    \begin{array}{cc}
        \boldsymbol{C} & \boldsymbol{0} \\
        \boldsymbol{0} & 2 \boldsymbol{C} \otimes \boldsymbol{C}
    \end{array}
    \right] ,
\end{align*}
where $\otimes$ is the Kronecker product. Hence, the natural gradient of the log-likelihood of $p(\btheta)$ is calculated as follows:
\begin{align}
    \ngr_\btheta \ln p(\btheta) = \left[
    \begin{array}{c}
        -\gamma (\boldsymbol{m} - \boldsymbol{\delta}) \\
        \VEC \left( \gamma (\boldsymbol{m} - \boldsymbol{\delta}) (\boldsymbol{m} - \boldsymbol{\delta})^\top + \PSI - (\nu + N + 2) \boldsymbol{C} \right)
    \end{array}
    \right]
\end{align}

\subsection{Update Rules for MAP-IGO with Multivariate Normal Distribution}
From \cite{Akimoto:2010}, the natural gradient of the log-likelihood of $p_\btheta (\x)$ is calculated as
\begin{align}
    \ngr_\btheta \ln p_\btheta (\x) = \left[
    \begin{array}{c}
        (\x - \boldsymbol{m}) \\
        \VEC \left( (\x - \boldsymbol{m})(\x - \boldsymbol{m})^\top - \boldsymbol{C} \right)
    \end{array}
    \right] \enspace.
\end{align}
% Let $\w_i = \hat{\w}_i / (\sum_{j=1}^\lambda \hat{\w}_j )$ for short, and we obtain the extended rank-$\mu$ update rule from Eq.~\eqref{eq:MAP-IGO-update}.
Let $\w_i = \hat{\w}_i / (\sum_{j=1}^\lambda \hat{\w}_j )$ for short, and we obtain the update rule from Eq.~\eqref{eq:MAP-IGO-update}.
\begin{align}
    \m[t+1] &= \m[t] + c_m \left( \sum_{i=1}^\lambda \w_i (\x_{i:\lambda} - \m[t] ) \highlight{- \gamma (\m[t] - \boldsymbol{\delta}) }  \right) \\
    \begin{split}
        \C[t+1] &= \C[t] + c_\mu \Biggl( \sum_{i=1}^\lambda \w_i \left( (\x_{i:\lambda} - \m[t])(x_{i:\lambda} - \m[t])^\top - \C[t] \right) \Biggr. \\
        &\qquad\qquad\qquad \Biggl. \highlight{ + \gamma (\m[t] - \boldsymbol{\delta}) (\m[t] - \boldsymbol{\delta})^\top \!+ \boldsymbol{\Psi} - (\nu + N + 2)\C[t] \hspace{-1pt} } \! \Biggr)
    \end{split} \label{eq:map-igo-C}
\end{align}
% Then, the MAP-CMA is obtained by combining this extended rank-$\mu$ update, the rank-one update, and the step-size adaptation.
% The single update in the MAP-IGO with multivariate normal distribution is shown in Algorithm~\ref{alg:map-igo}.
\colorbox[gray]{0.90}{Shaded terms} are differences from the original pure rank-$\mu$ update and represent terms corresponding to the natural gradient of the log-likelihood of the prior distribution.
% When the parameters of the prior distribution are set so that these natural gradient terms are zero vector or zero matrix, i.e., $\boldsymbol{\delta} = \m[t]$, $\PSI = (\nu + N + 2)\C[t]$, the MAP-CMA coincides with the original CMA-ES.

\begin{comment}
\begin{algorithm}[t]
    \caption{Single update in MAP-IGO with multivariate normal}
    \label{alg:map-igo}
    \begin{algorithmic}[1]
        \STATE \textbf{given } $\m[t] \in \R^N$, $\C[t] \in \R^{N \times N}$, $\ps[t] \in \R^N$, $\pc[t] \in \R^N$, $\gamma \in \R$, $\boldsymbol{\delta} \in \R^N$, $\PSI \in \R^{N \times N}$, $\nu \in \R$
        \FOR {$i = 1, \ldots, \lambda$}
            \STATE $\x_i \sim \mathcal{N}(\m[t], \C[t])$
        \ENDFOR
        \STATE $\m[t+1] \leftarrow \m[t] + c_{m} \left( \sum_{i=1}^{\lambda} w_{i} ( \x_{i:\lambda} - \m[t] ) \highlight{- \gamma (\m[t] - \boldsymbol{\delta}) } \right)$
        \STATE $\C[t+1] \leftarrow \C[t] + c_\mu \Biggl( \sum_{i=1}^\lambda w_i \left(\y_{i:\lambda}\y_{i:\lambda}^\top - \C[t]\right)$\linebreak
        \null\qquad\qquad\qquad\qquad $\highlight{ + \gamma (\m[t] - \boldsymbol{\delta}) (\m[t] - \boldsymbol{\delta})^\top + \Psi - (\nu + N + 2)\C[t] \hspace{-1pt} } \Biggr)$
    \end{algorithmic}
\end{algorithm}
\end{comment}
 % \label{sec:mapigo}
% \input{02_body/04_mapcma} % \label{sec:mapcma}
% ========================================
\section{Interpretation of the Rank-one Update with Prior Distribution} \label{sec:rankone}
% ========================================
In the previous section, we introduced the MAP-IGO to incorporate prior distribution into the IGO framework, and derived the update rules of the MAP-IGO in which the multivariate normal distribution and the normal-inverse-Wishart distribution are applied.
In this section, we demonstrate that the rank-one update can be derived from the MAP-IGO by setting the prior distribution based on the idea that the promising mean vector should exist in the direction of the evolution path.
In addition, we discuss the interpretation to the parameter setting of the prior distribution used to derive the rank-one update.

\subsection{Derivation of the Rank-one Update}
To confirm that we can derive the rank-one update from the MAP-IGO and reproduce the update rule of the CMA-ES, we first introduce the step-size $\sig[t]$ into the update rule of the MAP-IGO. By replacing $\C[t]$ with $(\sig[t])^2\C[t]$, Eq.~\eqref{eq:map-igo-C} results in the following:
\begin{align}
    \begin{split}
        \C[t+1] &= \C[t] \!+ c_\mu \Biggl( \sum_{i=1}^\lambda \w_i \left( \left(\frac{\x_{i:\lambda} - \m[t]}{\sig[t]} \right)\!\biggl(\frac{\x_{i:\lambda} - \m[t]}{\sig[t]}\right)^\top \!\!- \C[t] \biggr) \Biggr. \\
        &\qquad\quad \! \Biggl. + \gamma \biggl( \frac{\m[t] - \boldsymbol{\delta}}{\sig[t]} \biggr) \!\biggl(\frac{\m[t] - \boldsymbol{\delta}}{\sig[t]} \biggr)^{\!\top} \!\!\!+ \frac{\boldsymbol{\Psi}}{(\sig[t])^2} - (\nu + N + 2)\C[t] \! \Biggr)
    \end{split} \label{eq:map-igo-C-sig}
\end{align}
% We note that $\y_{i:\lambda} = (\x_{i:\lambda} - \m[t])/\sig[t]$ as shown in Eq.~\eqref{eq:sample_y}.
Next, we set the parameters $\boldsymbol{\delta}, \gamma, \PSI$ of the normal-inverse-Wishart distribution.
% When the direction of the mean vector move is obtained as the evolution path $\pc[t+1]$, the mean vector can be considered to likely exist at $\m[t] + r \sig[t] \pc[t+1]$ with $r>0$.
When the expected update direction of the mean vector is obtained as the evolution path $\pc[t+1]$, the promising mean vector can be considered to exist at $\m[t] + r \sig[t] \pc[t+1]$ with $r>0$.
%Next, let $r > 0$ and $\m[t] + r \sig[t] \pc[t+1]$ be the expected destination of the mean vector.
In this situation, we can set $\boldsymbol{\delta} = \m[t] + r \sig[t] \pc[t+1]$, which results from the fact that the expected value of the mean vector in the normal-inverse-Wishart distribution is given by $\boldsymbol{\delta}$.
To match the coefficients in Eq.~\eqref{eq:cmaes-C} and Eq.~\eqref{eq:map-igo-C-sig}, we set
\begin{align}
    \gamma = \frac{c_1}{r^2 c_\mu} \qquad \text{and} \qquad \boldsymbol{\Psi} = \left(\nu + N + 2 - \frac{c_1}{c_\mu} \right) (\sig[t])^2 \C[t] \enspace.
\end{align}
Therefore, we obtain the update rule of the mean vector and covariance matrix.
\begin{align}
    \m[t+1] &= \m[t] + c_m \left( \sum_{i=1}^\lambda \w_i (\x_{i:\lambda} - \m[t] ) ~\highlight{+ ~\frac{c_1}{r c_\mu} \sig[t] \pc[t+1]} \right) \label{eq:MAP-rankone-m} \\
    % \begin{split}
    %     \C[t+1] &= \C[t] + c_\mu \Biggl( \sum_{i=1}^\lambda \w_i \left( \y_{i:\lambda} \y_{i:\lambda}^\top - \C[t] \right) \Biggr. \\
    %     &\qquad\qquad\qquad\qquad\qquad\qquad\qquad \Biggl. +\frac{c_1}{c_\mu} \pc[t+1] {\pc[t+1]}^\top - \frac{c_1}{c_\mu} \C[t] \Biggr)
    % \end{split} \label{eq:MAP-rankone-C}
    \C[t+1] &= \C[t] + \!c_\mu \Biggl( \sum_{i=1}^\lambda \w_i \!\left( \y_{i:\lambda} \y_{i:\lambda}^\top - \C[t] \!\right) \!+\frac{c_1}{c_\mu} \biggl( \pc[t+1] {\pc[t+1]}^\top \!\!- \C[t] \biggr) \!\!\Biggr) \label{eq:MAP-rankone-C}
\end{align}
\renewcommand{\baselinestretch}{1.7}
{\setlength{\algomargin}{0.01em}
\begin{algorithm}[t]
    \caption{Single update in the MAP-CMA}
    \label{alg:map-cma}
    \begin{algorithmic}[1]
        \STATE \textbf{given } $\m[t] \in \R^N$, $\sig[t] \in \R_{+}$, $\C[t] \in \R^{N \times N}$, $\ps[t] \in \R^N$, $\pc[t] \in \R^N$
        \FOR {$i = 1, \ldots, \lambda$}
            \STATE $\y_i \sim \mathcal{N}(\boldsymbol{0}, \C[t])$
            \STATE $\x_i \leftarrow \m[t] + \sig[t] \y_i$
        \ENDFOR
        \STATE $\ps[t+1] \leftarrow (1-c_\sigma)\ps[t] + \sqrt{c_\sigma(2-c_\sigma)\muw} {\C[t]}^{-\frac12} \sum_{i=1}^\lambda \w_i \y_{i:\lambda}$
        \STATE $\pc[t+1] \leftarrow (1-c_c) \pc[t] + \sqrt{c_c(2-c_c)\muw} \sum_{i=1}^\lambda \w_i \y_{i:\lambda}$
        \STATE $\m[t+1] \leftarrow \m[t] + c_{m} \left( \sum_{i=1}^{\lambda} w_{i} ( \x_{i:\lambda} - \m[t] ) \highlight{+ \frac{c_1}{r c_\mu} \sig[t] \pc[t+1] \hspace{-2pt} } \right)$
        % \STATE $\m[t+1] \leftarrow \m[t] + c_{m} \left( \sum_{i=1}^{\lambda} w_{i} ( \x_{i:\lambda} - \m[t] ) + \frac{c_1}{r c_\mu} \sig[t] \pc[t+1] \right)$
        \STATE $\C[t+1] \leftarrow \C[t] + c_1 \left(\pc[t+1]{\pc[t+1]}^{\!\top} \!- \C[t]  \right) + c_\mu \sum_{i=1}^{\lambda} w_i \left( \y_{i:\lambda}\y_{i:\lambda}^\top - \C[t] \right)$
        \STATE $\sig[t+1] \leftarrow \sig[t] \exp \left( \frac{c_\sigma}{d_\sigma} \left( \frac{\|\ps[t+1] \|}{\E [\|\mathcal{N}(\boldsymbol{0}, \mathbf{I}) \|]} - 1 \right) \right)$
    \end{algorithmic}
\end{algorithm}
}
\renewcommand{\baselinestretch}{1.0}
We note that $\y_{i:\lambda} = (\x_{i:\lambda} - \m[t])/\sig[t]$ as defined in Eq.~\eqref{eq:sample_y}.
While the rank-one update of the covariance matrix is reproduced, the term \colorbox[gray]{0.90}{$\frac{c_m c_1}{r c_\mu} \sig[t] \pc[t+1]$} appears in the mean vector update rule.
As $r$ approaches infinity, Eq.~\eqref{eq:MAP-rankone-m} converges to the original update rule of CMA-ES shown in Eq.~\eqref{eq:cmaes-m}, whose interpretation is discussed in Section~\ref{ssec:interpretation2}.
When $r$ takes a finite value, the term \colorbox[gray]{0.90}{$\frac{c_m c_1}{r c_\mu} \sig[t] \pc[t+1]$} can be interpreted as a kind of \emph{momentum}, which is used in gradient descent methods.
We call this additional term \emph{momentum update} in this study.
In fact, Li and Zhang~\cite{Evopath:2016} show that the evolution path accumulates natural gradients with respect to the mean vector and acts as a momentum under stationary condition. They also show that the outer product of the evolution path serves as a rank-one momentum term for the covariance matrix.
% Therefore, our interpretation suggests that the rank-one update with the momentum term in the mean vector update is more natural than the original rank-one update.
Our interpretation suggests that the rank-one update with the momentum update is more rational than the original rank-one update in that it is derived in terms of natural gradients.
In addition, it can be said that the momentum update is conducive to achieving the goal of the rank-one update, which is to generate a solution in the direction of the evolution path.

Algorithm~\ref{alg:map-cma} shows the single-update procedure of the CMA-ES with the momentum update, named maximum a posterior CMA-ES (MAP-CMA). In this study, the MAP-CMA employs the cumulative step-size adaptation, which differs from the CMA-ES only by the momentum update, the shaded term.
The effect of the momentum update with finite $r$ is investigated in Section~\ref{sec:experiments}.

\subsection{Interpretation for the Setting of the Prior Distribution} \label{ssec:interpretation2}
\begin{figure}[t]
    \begin{center}
      \includegraphics[width=0.99\linewidth]{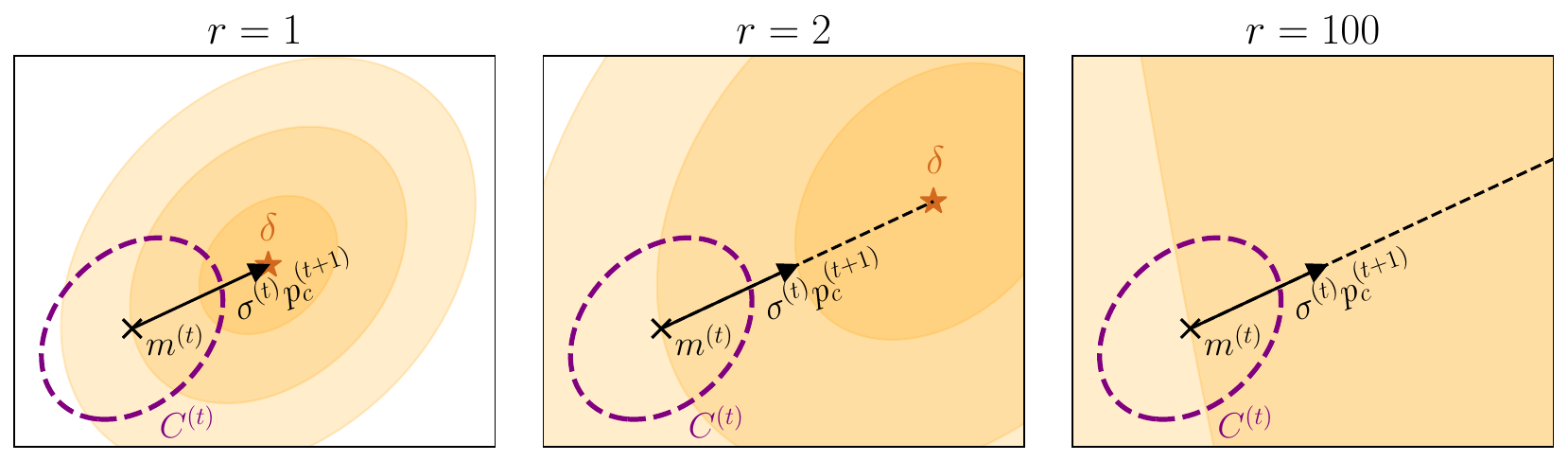}
    \end{center}
    \vspace{-2.0mm}
    \caption{The prior distribution with respect to the mean vector $\mathcal{N} \left( \boldsymbol{m} \mid \boldsymbol{\delta}, \frac{1}{\gamma} \boldsymbol{C} \right)$, where $\boldsymbol{\delta}$ and $\frac{1}{\gamma} \boldsymbol{C}$ are indicated by the orange star and ellipse, respectively. Since $\boldsymbol{\delta} = \m[t] + r \sig[t] \pc[t+1]$ and $\frac{1}{\gamma} \propto r^2$, multiplying $r$ by a constant $r'>1$ corresponds to expanding $\mathcal{N} \left( \boldsymbol{m} \mid \boldsymbol{\delta}, \frac{1}{\gamma} \boldsymbol{C} \right)$ by $r'$ times around $\m[t]$.}
    \label{fig:prior}
\end{figure}
In the normal-inverse-Wishart distribution, $\boldsymbol{\delta}$ and $\gamma$ are the expected value and global variance of the mean vector, respectively.
Multiplying $r$ by a constant $r' > 1$ corresponds to the affine transformation of expanding the prior distribution with respect to the mean vector by $r'$ times around $\m[t]$ as shown in Figure~\ref{fig:prior}.
% Thus, as $r$ increases, the prior distribution with respect to the mean vector $\mathcal{N} \left( \boldsymbol{m} \mid \boldsymbol{\delta}, \frac{1}{\gamma} \boldsymbol{C} \right)$ approaches a uniform distribution, i.e., a non-informative prior distribution.
Thus, as $r$ increases, the prior distribution with respect to the mean vector $\mathcal{N} \left( \boldsymbol{m} \mid \boldsymbol{\delta}, \frac{1}{\gamma} \boldsymbol{C} \right)$ approaches a non-informative prior distribution.
Then, as $r$ increases, the effect of the prior distribution on the mean vector update decreases. % \label{sec:rankone}
% ========================================
\section{Experiments} \label{sec:experiments}
% ========================================
In the previous section, we derived the rank-one update by setting the prior distribution.
Moreover, we showed that it is extensible by setting $r$ to a finite value and proposed MAP-CMA, a CMA-ES equipped with it.
In this section, we investigate the behavior of the MAP-CMA varying $r$.
% The code of MAP-CMA is available at \url{https://drive.google.com/drive/folders/1PQ8tO1BoX1omPpTvJL-e2R1Ae7QlSsXw?usp=sharing}\footnote{We will publicize our code upon publication.}.
The code of MAP-CMA will be made available at \textcolor{blue}{\url{https://github.com/CyberAgentAILab/cmaes}}~\cite{nomura2024cmaes}.

\begin{table*}[t]
    \caption{Benchmark functions to be minimized.}
    \label{table:benchmark}
    \centering
    \renewcommand{\arraystretch}{2.0}
    \begin{tabular}{ll}
        \hline
        Name & Definition \\
        \hline
        Sphere & $f(\x) = \sum_{i=1}^N x_i^2$ \\
        Ellipsoid & $f(\x) = \sum_{i=1}^N 10^{6\frac{i-1}{N-1}} x_i^2$ \\
        Cigar & $f(\x) = x_1^2 + \sum_{i=2}^N 10^6 x_i^2$ \\
        Rosenbrock & $f(\x) = \sum_{i=1}^{N-1} \Bigl(100(x_i^2 - x_{i+1})^2 + (x_i - 1)^2 \Bigr)$ \\
        Ackley & $f(\x) = 20 \!- \!20\exp \!\left( \!\!- 0.2 \sqrt{ \!\frac{1}{N} \!\sum_{i=1}^N \!x_i^2} \right) \!\!+ \!e \!- \!\exp \!\left( \frac{1}{N} \sum_{i=1}^N \cos (2\pi x_i) \right)$ \\
        Rastrigin & $f(\x) = 10N + \sum_{i=1}^N \Bigl( x_i^2 - 10 \cos(2\pi x_i) \Bigr)$ \\
        \hline
    \end{tabular}
    \renewcommand{\arraystretch}{1.0}
\end{table*}

%$\begin{aligned}f(\x) = 20 - 20\exp \left( - 0.2 \sqrt{\frac{1}{N} \sum_{i=1}^N x_i^2} \right) \\ + e - \exp \left( \frac{1}{N} \sum_{i=1}^N \cos (2\pi x_i) \right)\end{aligned}$ \\

\subsection{Experimental Setting}
The benchmark functions are summarized in Table~\ref{table:benchmark}.
\begin{comment}
We used following benchmark functions:
\begin{itemize}
\setlength{\itemsep}{3mm}
\setlength{\leftskip}{-2pt}
\item Sphere : $\sum_{i=1}^N x_i^2$
\item Ellipsoid : $\sum_{i=1}^N 10^{6\frac{i-1}{N-1}} x_i^2$
\item Cigar : $x_1^2 + \sum_{i=2}^N 10^6 x_i^2$
\item Rosenbrock : $\sum_{i=1}^{N-1} (100(x_i^2 - x_{i+1})^2 + (x_i - 1)^2)$
\item Ackley : $20 - 20\exp \left( - 0.2 \sqrt{\frac{1}{N} \sum_{i=1}^N x_i^2} \right) + e - \exp \left( \frac{1}{N} \sum_{i=1}^N \cos (2\pi x_i)
 \right)$
\item Rastrigin : $10N + \sum_{i=1}^N( x_i^2 - 10 \cos(2\pi x_i))$
\end{itemize}
\end{comment}
\begin{comment}
\begin{itemize}
\setlength{\itemsep}{3mm}
\setlength{\leftskip}{-4pt}
\item Sphere : $f(x) = \sum_{i=1}^N x_i^2$
\item Ellipsoid : $f(x) = \sum_{i=1}^N 10^{6\frac{i-1}{N-1}} x_i^2$
\item Ciagr : $f(x) = x_1^2 + \sum_{i=2}^N 10^6 x_i^2$
\item Rosenbrock : $f(x) = \sum_{i=1}^{N-1} (100(x_i^2 - x_{i+1})^2 + (x_i - 1)^2)$
\item Ackley : $f(x) = 20 - 20\exp \left( \! - 0.2 \sqrt{\frac{1}{N} \sum_{i=1}^N x_i^2} \right) + e - \exp \left( \frac{1}{N} \sum_{i=1}^N \cos (2\pi x_i)
 \right)$
\end{itemize}
\end{comment}
The initial mean vector $\m[0]$ was drawn uniform randomly from $[a, b]^N$ for each trial, and the initial step-size $\sig[0]$ was given by $(b-a)/2$, where $(a, b)$ was given by $(1, 5)$ for Sphere, Ellipsoid, Cigar, and Rastrigin, $(-2, 2)$ for Rosenbrock, $(1, 30)$ for Ackley.
The initial covariance matrix $\C[0]$ was given by an identity matrix.

\renewcommand{\arraystretch}{1.2}
\begin{table}[!t]
\centering
\caption{Success rate (SR) and SP1 over 100 independent trials.}
\begin{tabular}{|c|c|cc|cccccc|}
\hline
\multirow{3}{*}{\ds[Function]} & \multirow{3}{*}{$N$} & \multicolumn{2}{c|}{\multirow{2}{*}{CMA-ES}} & \multicolumn{6}{c|}{MAP-CMA}                                                       \\ \cline{5-10} 
                  &                   & \multicolumn{2}{c|}{}                  & \multicolumn{2}{c|}{$r=1$}    & \multicolumn{2}{c|}{$r=\sqrt{N}$}    & \multicolumn{2}{c|}{$r=N$} \\ \cline{3-10} 
                  &                   & SR & SP1 & SR & \multicolumn{1}{c|}{SP1} & SR & \multicolumn{1}{c|}{SP1} & SR & SP1 \\ \hline
\multirow{4}{*}{\ds[Sphere]} & \ds[10] & \ds[1.00] & \ds[{\bf 1787}] & \ds[1.00] & \multicolumn{1}{c|}{\ds[2399]\hspace{2pt}} & \ds[1.00] & \multicolumn{1}{c|}{\ds[1961]\hspace{2pt}} & \ds[1.00] & \ds[1824] \\
                  & \ds[20] & \ds[1.00] & \ds[{\bf 3329}] & \ds[1.00] & \multicolumn{1}{c|}{\ds[5034]\hspace{4pt}} & \ds[1.00] & \multicolumn{1}{c|}{\ds[3617]\hspace{4pt}} & \ds[1.00] & \ds[3366] \\
                  & \ds[40] & \ds[1.00] & \ds[{\bf 6276}] & \ds[1.00] & \multicolumn{1}{c|}{\ds[10719]\hspace{4pt}} & \ds[1.00] & \multicolumn{1}{c|}{\ds[6705]\hspace{4pt}} & \ds[1.00] & \ds[6322] \\
                  & \ds[80] & \ds[1.00] & \ds[{\bf 11568}] & \ds[1.00] & \multicolumn{1}{c|}{\ds[24518]\hspace{4pt}} & \ds[1.00] & \multicolumn{1}{c|}{\ds[12246]\hspace{4pt}} & \ds[1.00] & \ds[11658] \\ \hline
\multirow{4}{*}{\ds[Ellipsoid]} & \ds[10] & \ds[1.00] & \ds[6078] & \ds[1.00] & \multicolumn{1}{c|}{\ds[6888]\hspace{4pt}} & \ds[1.00] & \multicolumn{1}{c|}{\ds[6215]\hspace{4pt}} & \ds[1.00] & \ds[{\bf 6050}] \\
                  & \ds[20] & \ds[1.00] & \ds[18906] & \ds[1.00] & \multicolumn{1}{c|}{\ds[21486]\hspace{4pt}} & \ds[1.00] & \multicolumn{1}{c|}{\ds[18984]\hspace{4pt}} & \ds[1.00] & \ds[{\bf 18853}] \\
                  & \ds[40] & \ds[1.00] & \ds[68401] & \ds[1.00] & \multicolumn{1}{c|}{\ds[74256]\hspace{4pt}} & \ds[1.00] & \multicolumn{1}{c|}{\ds[{\bf 67205}]\hspace{4pt}} & \ds[1.00] & \ds[68056] \\
                  & \ds[80] & \ds[1.00] & \ds[265855] & \ds[1.00] & \multicolumn{1}{c|}{\ds[284614]\hspace{4pt}} & \ds[1.00] & \multicolumn{1}{c|}{\ds[{\bf 259434}]\hspace{4pt}} & \ds[1.00] & \ds[264651] \\ \hline
\multirow{4}{*}{\ds[Cigar]} & \ds[10] & \ds[1.00] & \ds[{\bf 4423}] & \ds[1.00] & \multicolumn{1}{c|}{\ds[5623]\hspace{4pt}} & \ds[1.00] & \multicolumn{1}{c|}{\ds[4768]\hspace{4pt}} & \ds[1.00] & \ds[4506] \\
                  & \ds[20] & \ds[1.00] & \ds[{\bf 8693}] & \ds[1.00] & \multicolumn{1}{c|}{\ds[12314]\hspace{4pt}} & \ds[1.00] & \multicolumn{1}{c|}{\ds[9256]\hspace{4pt}} & \ds[1.00] & \ds[8787] \\
                  & \ds[40] & \ds[1.00] & \ds[{\bf 16976}] & \ds[1.00] & \multicolumn{1}{c|}{\ds[26003]\hspace{4pt}} & \ds[1.00] & \multicolumn{1}{c|}{\ds[17738]\hspace{4pt}} & \ds[1.00] & \ds[17068] \\
                  & \ds[80] & \ds[1.00] & \ds[{\bf 31951}] & \ds[1.00] & \multicolumn{1}{c|}{\ds[58000]\hspace{4pt}} & \ds[1.00] & \multicolumn{1}{c|}{\ds[33227]\hspace{4pt}} & \ds[1.00] & \ds[32098] \\ \hline
\multirow{4}{*}{\ds[Rosenbrock]} & \ds[10] & \ds[0.93] & \ds[6972] & \ds[0.91] & \multicolumn{1}{c|}{\ds[7039]\hspace{4pt}} & \ds[0.91] & \multicolumn{1}{c|}{\ds[6825]\hspace{4pt}} & \ds[0.94] & \ds[{\bf 6802}] \\
                  & \ds[20] & \ds[0.90] & \ds[24015] & \ds[0.80] & \multicolumn{1}{c|}{\ds[24152]\hspace{4pt}} & \ds[0.92] & \multicolumn{1}{c|}{\ds[{\bf 20599}]\hspace{4pt}} & \ds[0.91] & \ds[22824] \\
                  & \ds[40] & \ds[0.93] & \ds[87489] & \ds[0.76] & \multicolumn{1}{c|}{\ds[80442]\hspace{4pt}} & \ds[0.89] & \multicolumn{1}{c|}{\ds[{\bf 77235}]\hspace{4pt}} & \ds[0.94] & \ds[82818] \\
                  & \ds[80] & \ds[0.91] & \ds[354686] & \ds[0.80] & \multicolumn{1}{c|}{\ds[{\bf 258692}]\hspace{4pt}} & \ds[0.92] & \multicolumn{1}{c|}{\ds[ 287489]\hspace{4pt}} & \ds[0.91] & \ds[344089] \\ \hline
\multirow{4}{*}{\ds[Ackley]} & \ds[10] & \ds[0.97] & \ds[{\bf 3719}] & \ds[0.73] & \multicolumn{1}{c|}{\ds[6734]\hspace{4pt}} & \ds[0.94] & \multicolumn{1}{c|}{\ds[4248]\hspace{4pt}} & \ds[0.96] & \ds[3964] \\
                  & \ds[20] & \ds[0.95] & \ds[{\bf 6946}] & \ds[0.60] & \multicolumn{1}{c|}{\ds[17122]\hspace{4pt}} & \ds[0.96] & \multicolumn{1}{c|}{\ds[7555]\hspace{4pt}} & \ds[0.96] & \ds[6991] \\
                  & \ds[40] & \ds[1.00] & \ds[{\bf 12143}] & \ds[0.31] & \multicolumn{1}{c|}{\ds[69112]\hspace{4pt}} & \ds[0.98] & \multicolumn{1}{c|}{\ds[13318]\hspace{4pt}} & \ds[1.00] & \ds[12316] \\
                  & \ds[80] & \ds[1.00] & \ds[{\bf 21974}] & \ds[0.00] & \multicolumn{1}{c|}{\ds[-]\hspace{4pt}} & \ds[0.91] & \multicolumn{1}{c|}{\ds[25775]\hspace{4pt}} & \ds[0.99] & \ds[22314] \\ \hline
\multirow{4}{*}{\ds[Rastrigin]} & \ds[10] & \ds[0.99] & \ds[50781] & \ds[1.00] & \multicolumn{1}{c|}{\ds[{\bf 50113}]\hspace{4pt}} & \ds[0.98] & \multicolumn{1}{c|}{\ds[50722]\hspace{4pt}} & \ds[0.97] & \ds[51602] \\
                  & \ds[20] & \ds[1.00] & \ds[{\bf 167412}] & \ds[1.00] & \multicolumn{1}{c|}{\ds[168168]\hspace{4pt}} & \ds[1.00] & \multicolumn{1}{c|}{\ds[167748]\hspace{4pt}} & \ds[1.00] & \ds[168196] \\
%                   & \ds[40] & \ds[0.98] & \ds[{\bf 512697}] & \ds[0.96] & \multicolumn{1}{c|}{\ds[528676]\hspace{4pt}} & \ds[0.95] & \multicolumn{1}{c|}{\ds[528417]\hspace{4pt}} & \ds[0.94] & \ds[537786] \\
                  & \ds[40] & \ds[1.00] & \ds[{\bf 606858}] & \ds[0.99] & \multicolumn{1}{c|}{\ds[620722]\hspace{4pt}} & \ds[1.00] & \multicolumn{1}{c|}{\ds[610302]\hspace{4pt}} & \ds[1.00] & \ds[608874] \\
                  & \ds[80] & \ds[1.00] & \ds[2110892] & \ds[1.00] & \multicolumn{1}{c|}{\ds[2127692]\hspace{4pt}} & \ds[1.00] & \multicolumn{1}{c|}{\ds[{\bf 2096612}]\hspace{4pt}} & \ds[1.00] & \ds[2106244] \\ \hline
\end{tabular}
\label{table:SRandSP1}
\end{table}

In the CMA-ES, $c_m$ was set to $1$. In the MAP-CMA, $c_m$ was set to $1/(1 + c_1/(c_\mu r))$ ensuring $c_m + c_m c_1 / (r c_\mu) = 1$.
The population size $\lambda$ of the CMA-ES and MAP-CMA was set to $4+ \lfloor 3 \ln N \rfloor$ for Sphere, Ellipsoid, Cigar, Rosenbrock, and Ackley, and for Rastrigin, it was set to $\{700, 1400, 2100, 2800\}$ for dimensions $\{10, 20, 40, 80\}$ so that the success rate for the CMA-ES was sufficiently high, referring to \cite{hansen2004evaluating}.
The other hyperparameters of the CMA-ES and MAP-CMA were set to those in \cite[Table~2]{CMAparam:2014}.

Each trial was terminated and regarded as a success if the best evaluation value reached less than $10^{-10}$.
Each trial was terminated and regarded as a failure if any of the following conditions were met: the number of evaluations reached more than $10^6 N$; the minimum eigenvalue of $(\sig[t])^2 \C[t]$ became less than $10^{-30}$.
For each setting, the 100 independent trials were conducted.

\subsection{Results and Discussion}
\begin{figure}[t]
    \begin{center}
      \includegraphics[width=0.99\linewidth]{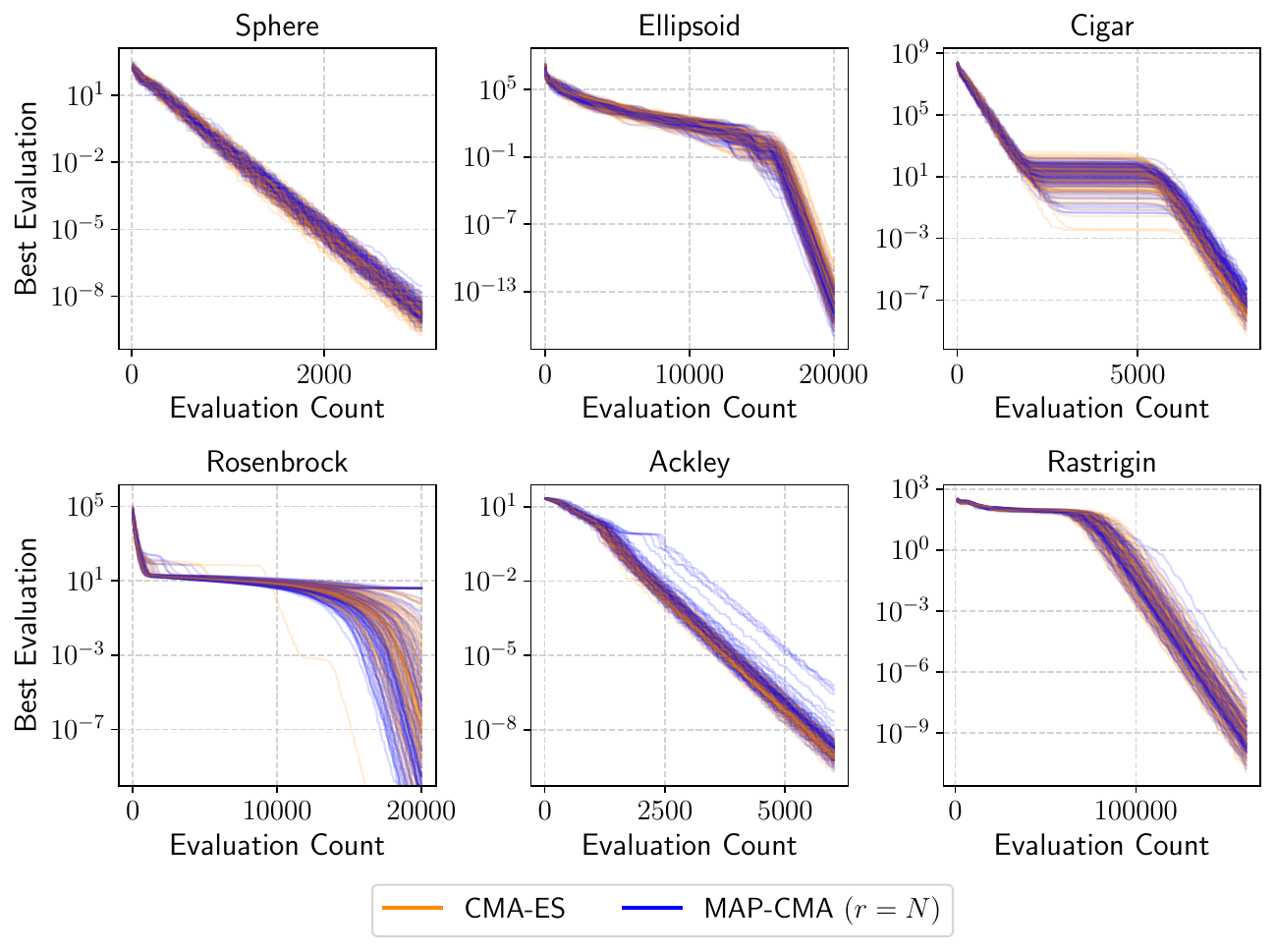}
    \end{center}
    \vspace{-2.0mm}
    \caption{Transitions of best evaluation value for $N=20$ over 100 independent trials.}
    \label{fig:transition}
\end{figure}

Table~\ref{table:SRandSP1} shows the success rate and SP1~\cite{auger2005restart} over 100 independent trials. We note that the SP1 index is defined as the average evaluation counts in successful trials divided by the success rate.
\del{The MAP-CMA with $r=1$ suffers from instability except for Sphere.
This is due to that the effect of the momentum update, i.e., the old samples accumulated in the evolution path, is too large.
In most cases, the MAP-CMA with $r=\sqrt{N}$ is marginally better than the CMA-ES on the SP1 index.}{}\new{The behavior of the MAP-CMA approaches that of the CMA-ES as $r$ increases. When $r$ is too small and the effect of the old samples accumulated in the evolution path is too large, the momentum update in the MAP-CMA tends to have a negative effect on optimization.
When $r=N$, the MAP-CMA is superior to the CMA-ES for Rosenbrock and competitive for the other functions.}
Figure~\ref{fig:transition} shows the transitions of the best evaluation value on the 100 independent trials.
\del{For Sphere, Cigar, Rosenbrock, and Ackley, the MAP-CMA improves the best evaluation values more efficiently than the CMA-ES.
This is probably because that the momentum update, which contains the evolution path in the mean vector update, reduces the variance of the update direction.}{}\new{While the difference in performance between the CMA-ES and the MAP-CMA is slight for Sphere, Ellipsoid, Cigar, and Rastrigin, the MAP-CMA improves the best evaluation values more efficiently than the CMA-ES for Rosenbrock.}{}

\begin{figure}[t]
    \begin{center}
      \includegraphics[width=0.99\linewidth]{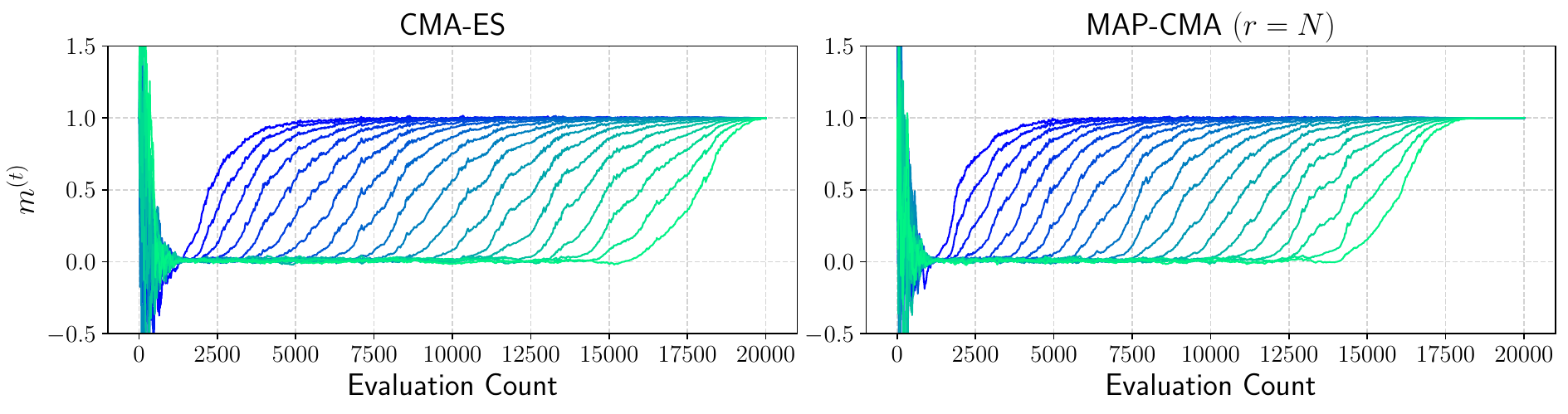}
    \end{center}
    \vspace{-2.0mm}
    \caption{Transition of mean vector in one typical trial of optimizing Rosenbrock with $N=20$.}
    \label{fig:rosenbrock}
\end{figure}

Figure~\ref{fig:rosenbrock} shows the transition of the mean vector when optimizing Rosenbrock with $N=20$.
The initial mean vector and step-size were given as $\m[0] = (1, \ldots, 1)$ and $\sig[0] = 1$, respectively.
Once the mean vector reaches near the origin, it moves to $(1, \ldots, 1)$.
At this stage, the momentum update in the mean vector update allows the MAP-CMA to move the mean vector faster than the CMA-ES.
\new{This result suggests that the MAP-CMA can be more efficient than the CMA-ES when the required mean vector moves are large.}
% This result corroborates that the MAP-CMA is more efficient than the CMA-ES when the required mean vector moves are large.

\del{For Ellipsoid optimization, the performance difference between the CMA-ES and MAP-CMA is small because covariance matrix adaptation is the bottleneck of the optimization.}{}When optimizing Ackley with the MAP-CMA, the optimization stalls in some trials.
This stagnation may be due to the effect of past samples~\cite{shirakawa2015sample}, which is an issue to be investigated in the future.
% The MAP-CMA is successful in most trials of Rastrigin, but for the same reason as Ackley, the performance difference with the CMA-ES disappears.
\del{While the MAP-CMA is mostly successful in Rastrigin, its advantage over the CMA-ES diminishes due to the effect of past samples, similar to the situation observed in Ackley.}{}
% These results suggest that there is room for improvement in the hyperparameters of the MAP-CMA.

 % \label{sec:experiments}
% ========================================
\section{Conclusion} \label{sec:conclusion}
% ========================================
We provided a new interpretation of the rank-one update in the CMA-ES from the perspective of the natural gradient with prior distribution.
We firstly proposed maximum a posteriori IGO (MAP-IGO), which is the IGO framework extended to incorporate a prior distribution, and derived the update rules of the MAP-IGO in which the multivariate normal distribution and the normal-inverse-Wishart distribution are applied. 
Then, we derived the rank-one update from the MAP-IGO by setting the prior distribution based on the idea that the promising mean vector should exist in the direction of the evolution path. 
Furthermore, the newly derived rank-one update is extensible, and we proposed maximum a posterior CMA-ES (MAP-CMA), a CMA-ES equipped with it.
The mean update rule in the MAP-CMA has an additional term containing the evolution path, which can be interpreted as a momentum term as discussed in~\cite{Evopath:2016} and we call this additional term \emph{momentum update} in this study.
% Numerical experiments showed that the MAP-CMA outperforms the CMA-ES on various benchmark functions.
Experimental results showed that \del{the MAP-CMA is marginally better than the CMA-ES in optimizing the various benchmark functions.}{}\new{the MAP-CMA is better than the CMA-ES in optimizing functions that require a lot of mean vector moves, such as Rosenbrock.}

In this study, the coefficients of the prior distribution were set so that the derived rank-one update matched that of the original CMA-ES.
% However, these coefficients and the current hyperparameters of the MAP-CMA are not necessarily optimal because some trials stagnate in the multimodal function.
However, there is room for adjustment of these coefficients and the current hyperparameters of the MAP-CMA because some trials stagnate in the multimodal function.
The detailed investigation of these settings is future work.
In addition, it is important to research the effectiveness of momentum update in the CMA-ES with restart strategies~\cite{IPOP:2005, BIPOP:2009} or integer handling~\cite{CMAESIM:2011, CMAESwM:2022}.
% Additionally, it is important to research the effectiveness of momentum update in other variants of CMA-ES, such as CMA-ES with Margin.
Furthermore, the MAP-IGO can set the parameter of the prior distribution other than the idea of the rank-one update, and it is also possible to derive the PBIL that can handle the prior distribution by applying the Bernoulli and Beta distributions.
The discovery of new knowledge and the improvement of algorithms based on these techniques are also important future works.
 % \label{sec:conclusion}

\bibliographystyle{splncs04}
\bibliography{ref}

\end{document}